\documentclass[11pt]{article}

\usepackage{ACL2023}

\usepackage{times}
\usepackage{latexsym}

\usepackage[T1]{fontenc}

\usepackage[utf8]{inputenc}

\usepackage{microtype}

\usepackage{inconsolata}

\usepackage{graphicx}
\usepackage{booktabs}       
\usepackage{algorithm}
\usepackage{algpseudocode}
\usepackage{tabularx, makecell, multirow}
\title{Improving Domain Adaptation through Extended-Text Reading Comprehension}

\author{
\textbf{Ting Jiang}\textsuperscript{\rm 1} , \textbf{Shaohan Huang}\textsuperscript{\rm 2} , \textbf{Shengyue Luo} \textsuperscript{\rm 2}, \textbf{Zihan Zhang}\textsuperscript{\rm 2} , \textbf{Haizhen Huang}\textsuperscript{\rm 2}\\
\textbf{Furu Wei}\textsuperscript{\rm 2} , \textbf{Weiwei Deng}\textsuperscript{\rm 2} , \textbf{Feng Sun}\textsuperscript{\rm 2} , \textbf{Qi Zhang}\textsuperscript{\rm 2} , \textbf{Deqing Wang$^\dagger$}\textsuperscript{\rm 1} , \textbf{Fuzhen Zhuang}\textsuperscript{\rm 1} \\
\textsuperscript{\rm 1}Beihang University \textsuperscript{\rm 2}Microsoft Corporation\\
\texttt{royokong@buaa.edu.cn}\\
}
\begin{document}
\maketitle

\begin{abstract}
To enhance the domain-specific capabilities of large language models, continued pre-training on a domain-specific corpus is a prevalent method. Recent work demonstrates that adapting models using reading comprehension data formatted by regex-based patterns can significantly improve performance on domain-specific tasks.
However, regex-based patterns are incapable of parsing raw corpora using domain-specific knowledge. Furthermore, the question and answer pairs are extracted directly from the corpus in predefined formats offers limited context.
To address this limitation, we improve reading comprehension via LLM and clustering.
LLM focuses on leveraging domain knowledge within the corpus to refine comprehension stage,
while clustering supplies relevant knowledge by extending the context to enrich reading stage.
 Additionally, our method incorporates parameter-efficient fine-tuning to improve the efficiency of domain adaptation.
In comparison to AdaptLLM, our method achieves an improvement exceeding 5\% in domain-specific tasks.
Our code will available at \url{https://github.com/microsoft/LMOps}.

\end{abstract}

\section{Introduction}

With the emergence of Large Language Models (LLMs), LLMs have shown promising performance on various downstream tasks.
A number of domain-specific LLMs~\cite{cheng2023adapting, chen2023disc, wu2023bloomberggpt, han2023medalpaca, liu2023chipnemo} have also been proposed to enhance LLMs on domain-specific capabilities of LLMs, which demonstrate improved performances in respective domains compared to general models.
These domain-specific LLMs can be trained in two ways: either from scratch or by adapting existing general LLMs through continued pre-training~\cite{gururangan2020don}, with the latter being a more efficient method due to the foundational benefits provided by the general LLMs.

Recent work~\cite{cheng2023adapting} reveals that straightforward adaptation of a general LLM using raw domain-specific corpus is ineffective and can even impair prompting ability on domain-specific tasks.
To harness the potential of domain-specific knowledge, they proposed a data preprocessing method named AdaptLLM. This method transforms a corpus into a reading comprehension format, utilizing specially designed regex-based patterns.
Consequently, AdaptLLM notably enhances the performance of domain-specific tasks by structuring a corpus in the question-answering format.

However, the reliance on regex-based patterns poses challenges in handling complex patterns and generating questions that reflect domain-specific knowledge.
For example, the regex-based pattern \texttt{\{SENT1\} Therefore, \{SENT2\}} is converted into a question-answer format as: \texttt{What is the cause of \{SENT1\}? \{SENT2\}}. This method also limits the diversity of question types.
Integrating LLMs in the preprocessing stage can overcome these limitations. LLMs like ChatGPT are capable of identifying domain-specific knowledge and generating high-quality question-answer pairs for educational purposes~\cite{olney2023generating, lu2023can}.
To mitigate the processing costs associated with ChatGPT in preprocessing, we fine-tune a compact LLM through knowledge distillation, to efficiently preprocesses domain-specific data.

\begin{figure*}
	\includegraphics[width=2\columnwidth]{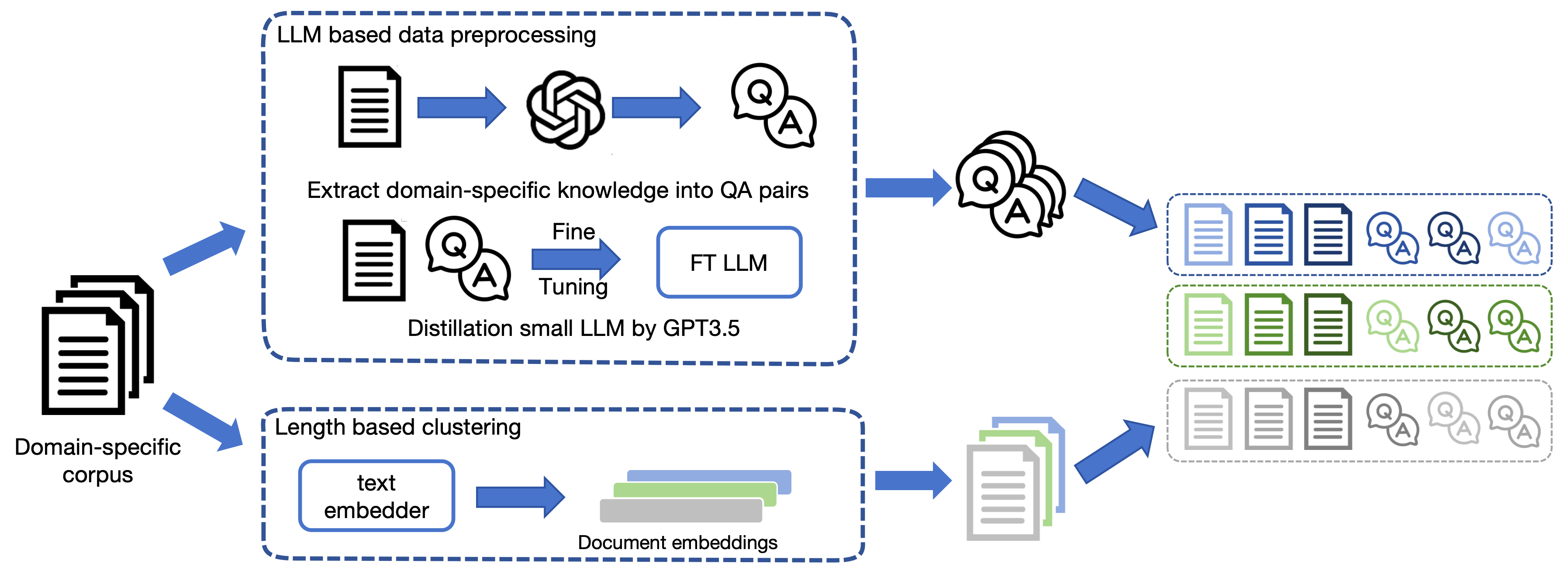}
\caption{
  The overall framework of our method.
  Best view in color.
} \label{fig:framework}
\end{figure*}
We find that the context of question answering can be too short to learn domain-specific knowledge comprehensively. For example, in biomedicine, each document is a short abstract of paper, which is easy for LLM to answer questions, but lacks enough context to learn domain-specific knowledge.
Inspired by~\cite{levine2021inductive, shi2023context}, we leverage length-based clustering to extend the context by concatenating similar documents into the same input as context.
Moreover, we improve the efficiency of domain adaptation by utilizing parameter-efficient fine-tuning methods like LoRA~\cite{hu2021lora}.
Contrary to previous work~\cite{liu2023chipnemo}, we find that LoRA can be more efficient than full fine-tuning for domain adaptation with proper settings.

\section{Methods}
Considering the constraints of AdaptLLM in converting the corpus into reading comprehension via regex-based patterns, our method improves the quality of question answer pairs via LLM to enhance comprehension phase and extends their context by clustering to enhance reading phase, as illustrated in Figure~\ref{fig:framework}.
Furthermore, we employ parameter-efficient fine-tuning to enhance domain adaptation efficiency with appropriate settings.

\subsection{LLM-based data Preprocessing}

For question-answer pairs generation, we employ ChatGPT to generate question-answer pairs with the prompt template:
\texttt{\{DOCUMENT\} Ask a few questions to help understand the above passage about \{DOMAIN\} and give the corresponding answers in JSON list (each JSON contain two keys: question and answer).}
Here, \texttt{\{DOCUMENT\}} represents the text from the corpus such as the paper abstract in biomedicine domain. \texttt{\{DOMAIN\}} indicates the adapted domain, which could be biomedicine or finance.

Considering that the corpus can contain more than a billion tokens, preprocessing the entire domain specific corpus can be expensive with the API based LLMs.
To solve this problem, we further fine-tune a 7B parameter LLM by distilling from ChatGPT to generate question-answer pairs for entire corpus.

\subsection{Length-based Clustering}
We leverage document similarity to extend the context of question answering.
Specifically, we embed documents with the text embedding model to cluster documents.
Since the document length varies, we stop adding a new document to the cluster when the length exceeds the threshold or achieves the maximum amount of documents.
To format the text of a cluster, we first concatenate all the documents, then shuffle their question-answer pairs to assemble them.
Following AdaptLLM, we also augment the domain corpus with general instructions.
Finally, we use 0/1 knapsack algorithm to fit these into the maximum context length of LLMs.
The detailed algorithm is in Algorithm~\ref{alg:clustering}.

\begin{table*}[t]
\centering
\begin{tabular}{lcccccc}
\toprule
\midrule
\multicolumn{7}{c}{\it{Biomedicine}}\\
\midrule
                             & \textbf{BioMMLU} & \textbf{PubMedQA} &  \textbf{MQP} & \textbf{RCT} & \textbf{UMSLE} & \textbf{Avg.} \\
\midrule
General LLM               & 29.9             & 74.0              & 50.0          & 47.0         & 28.9           & 46.0  \\
AdaptLLM$^\dagger$ & 46.6             & 75.2              & 68.7          & 47.1         & 33.3           & 54.2 \\
\midrule
DAPT                 & 47.2             & 73.6              & 50.8          & 32.3         & 38.7           & 48.5  \\
ReadCompre & 47.3 & \textbf{75.2} & 68.8 & 47.0 & 39.8 & 55.6 \\
Our & \textbf{48.3}             & 73.7              & \textbf{79.8}          & \textbf{58.0}         & \textbf{40.1}           & \textbf{60.0}  \\
\bottomrule
\midrule
\multicolumn{7}{c}{\it{Finance}}\\
\midrule
                                       & \textbf{FiQA SA} &\textbf{ConvFinQA} & \textbf{FPB} &\textbf{NER} & \textbf{Headline} & \textbf{Avg.}\\
\midrule
General LLM                         & 40.5             & 38.3             & 20.6         & 67.6        & 84.6              & 50.3 \\
AdaptLLM$^\dagger$   & 65.6             & \textbf{46.9}              & 58.1         & 69.1        & 85.7              & 65.1 \\
\midrule
DAPT & 75.6 & 42.2 & 67.4 & 73.1 & 85.2 & 68.7\\
ReadCompre & 75.7 & 39.9 & 70.9 & 68.4 & \textbf{86.8} & 68.3 \\
Our                                    & \textbf{77.3}             &  43.8             & \textbf{77.6}         & \textbf{70.3}        & 84.4              & \textbf{70.7} \\

\bottomrule
\end{tabular}
\caption{
  Main results on domain-specific task performance with general LLM, AdaptLLM~\cite{cheng2023adapting}, domain-adaptive pre-training (DAPT), data preprocessing following in AdaptLLM (ReadCompre) and our method.
  For DAPT and ReadCompre, we reproduce these methods with the same training setting and domain corpus to demonstrate the effectiveness of our method.
  \(\dagger\): results from evaluating the published checkpoints. We find the performance of AdaptLLM is under-estimated in the original paper due to the dirty data and messy templates, and re-evaluate the performance of AdaptLLM with cleaned data and format templates.
}\label{tab:main_results}
\vspace{-10pt}
\end{table*}

\begin{algorithm}
\small
\caption{Length-based Clustering}\label{alg:clustering}
\textbf{Input:} Set of documents $D$, Set of question-answer pairs $P$, Set of general instructions $G$,Text embedding model $M$, Length threshold $L_{max}$, Max documents $D_{max}$\\
\textbf{Output:} Model input $I$
\begin{algorithmic}[1]
\State Initialize clusters $C \gets \emptyset$
\For{each document $d \in D$}
    \State Embed $d$ with $M$ to get embedding $e_d$
\EndFor
\While{$|\mathcal{D}| > 0$}
    \State Randomly select $d_i$ from $D$ as cluster $c$
    \State Initialize length $L_c \gets len(d_i) + len(p_i)$
    \State Initialize count $N_c \gets 1$
    \While{$L_c < L_{max}$ and $N_c < D_{max}$}
        \State Get the closet $d_i$ to cluster $c$ based on $e_{d}$
        \If{$sim(d_i, c) < 0.7$}
        \State Break \Comment{stop clustering if no similar $d$}
        \EndIf
        \State Update $L_c$ and $N_c$ based on $d_i$ and $p_i$
    \EndWhile
    \State Remove $c$ from $D$ and append to $C$
\EndWhile
\State Format each cluster $c$ in $C$
\State Tokenize $I= C \bigcup G$ with LLM tokenizer
\State Using 0/1 knapsack algorithm to fit $I$ with maximum context length
\State \Return{$I$}
\end{algorithmic}
\end{algorithm}

\subsection{Parameter Efficient Domain Adaptation}
Recent work~\cite{liu2023chipnemo} indicates that Parameter Efficient Fine-Tuning (PEFT) methods like LoRA~\cite{hu2021lora} are generally less effective than full fine-tuning for domain adaptation.
This shortcoming is attributed to the exclusive implementation of LoRA in the self-attention layers, while neglecting the feed-forward layers which are related to storage knowledge in LLMs~\cite{dai2021knowledge}.
It limits the ability of LoRA to store domain-specific knowledge during continued fine-tuning.
Distinct from other downstream tasks, such as translation, domain adaptation also exhibits heightened sensitivity to the quantity of trainable parameters.
For instance, a low LoRA rank, such as 8, is often adequate for standard tasks. However, domain adaptation requires a significantly higher rank, like 256, to achieve comparable results with full fine-tuning.
Nonetheless, even at a rank of 256, LoRA maintains a significant efficiency advantage over full fine-tuning.

\begin{table}
\centering
\small
\begin{tabular}{lccc}
\toprule
& \multirowcell{2}[0pt][c]{BioMMLU}&BioMMLU& \multirowcell{2}[0pt][c]{Improv.} \\
& &RAG&  \\
\midrule
DAPT &  47.4 &      52.5 & 5.1 \\
ReadCompre &  47.3 &      51.5 & 4.2 \\
Our          &  48.3 &      54.5 & 6.2 \\
 \bottomrule
\end{tabular}
\caption{
  Retrieval Augmented Generation (RAG) results on BioMMLU.
  We use LLM-Embedder~\cite{zhang2023retrieve} as the text embedder with the MS MARCO~\cite{nguyen2016ms} as the retrieval corpus.
}\label{tag:rag}
\end{table}

\section{Experiments}

\subsection{Experiment Settings}
Following the AdaptLLM~\cite{cheng2023adapting}, we use PubMed abstracts from the Pile~\cite{gao2020pile} for biomedicine domain, and financial news collected by FinGPT~\cite{liu2023fingpt} for finance domain. Additionally, the LIMA~\cite{zhou2023lima}, WizardLM~\cite{xu2023wizardlm}, and Orca~\cite{mukherjee2023orca} datasets are used as general instruction datasets with same mixing ratio as AdaptLLM. For data preprocessing, \texttt{gpt-3.5-turbo} is employed to generate question-answer pairs for 10000 documents in each domain to fine tune a 7B LLaMA~\cite{touvron2023llama}. This fine-tuned model is then utilized to generate question-answer pairs for the entire corpus.
For continue training, we use LoRA with a rank of 256 and int8 quantization to improve the efficiency of training.

For domain-specific tasks, we evaluate model on the following datasets: PubMedQA~\cite{jin2019pubmedqa}, MQP~\cite{mccreery2020effective}, RCT~\cite{dernoncourt2017pubmed}, USMLE~\cite{jin2021disease}, BioMMLU, which we select biomedicine subjects from MMLU~\cite{hendrycks2020measuring}, and ConvFinQA~\cite{chen2022convfinqa}, FPB~\cite{malo2014good}, NER~\cite{salinas-alvarado-etal-2015-domain}, Headline~\cite{sinha2021impact}, FiQA SA~\cite{maia201818} for finance domain.

\subsection{Main Results}

We show the main results on domain-specific tasks in Table~\ref{tab:main_results}.
Our method surpasses AdaptLLM, yielding average improvement of 6.8\% in biomedicine and 5.6\% in finance.
We also reproduce the results of DAPT and ReadCompre using identical training data and parameter efficient fine-tuning.
In the case of ReadCompre, which uses the same data preprocessing method as AdaptLLM, we are able to reproduce the results and even achieve better performance.
Compared to it, our method still achieves consistently improvements under the same data and settings, which demonstrates the effectiveness of our method. We do not find DAPT has an adverse impact on the general LLM as suggested in~\cite{cheng2023adapting}. Instead, DAPT even demonstrates better performance in finance compared to ReadCompre.
Furthermore, the integration of clustering to extend the context also enhance the performance on Retrieval-Augmented Generation (RAG), as detailed in Table~\ref{tag:rag}.

\begin{figure}
\vspace{+3pt}
	\includegraphics[width=\columnwidth]{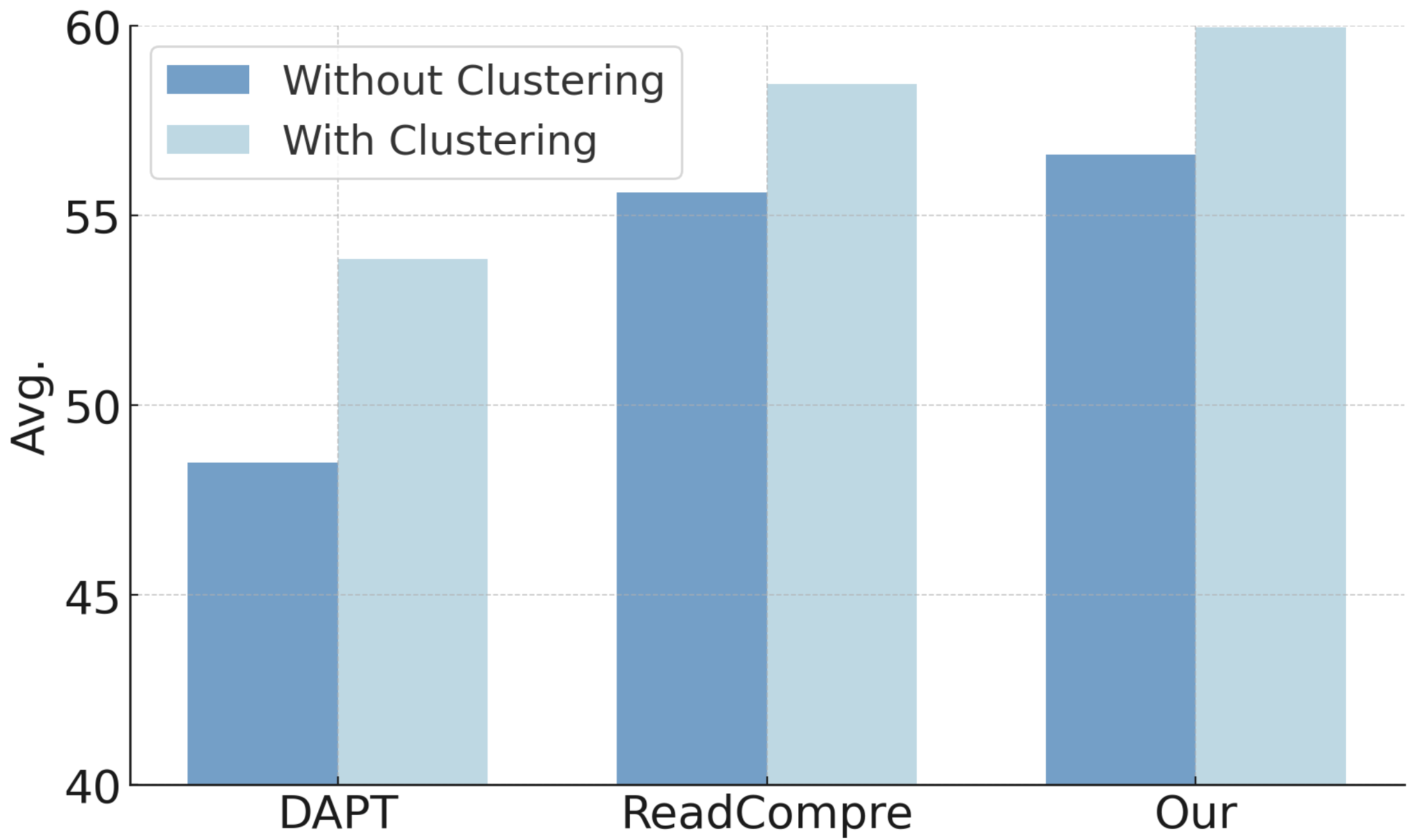}
\caption{
  Ablation study on clustering on biomedicine with DAPT, ReadCompre and our method.
}~\label{fig:clustering}
\end{figure}

\section{Ablation Study}
\subsection{Effect of Clustering}
To validate the effectiveness of our clustering method, we report the performance of DAPT, ReadCompre and our method with and without clustering  in Figure~\ref{fig:clustering}.
We find that clustering improves the performance of all three methods, which demonstrates the effectiveness of clustering in domain adaptation.

\begin{table}
\centering
\begin{tabular}{ccccc}
\toprule
 LoRA& LoRA & Trainable  & \multirowcell{2}[0pt][c]{Avg.} \\
 Rank & Target & Parameters & \\
\midrule
 \multicolumn{2}{c}{full fine-tuning} & 7B & 59.5 \\
\midrule
256 & QV linear & 256M & 53.6 \\
\midrule
8 & all linear  & 19M  & 50.1 \\
32 & all linear  & 153M &  52.5\\
128 & all linear  & 305M & 53.2 \\
256 & all linear & 610M & 60.0 \\
 \bottomrule
\end{tabular}
\caption{
  Ablation study on parameter efficient fine-tuning.
}\label{tab:ablation_lora}
\end{table}

\subsection{Effect of Parameter Efficient Fine-Tuning}
To study the influence of parameter efficient fine-tuning, we report the performance on biomedicine of different LoRA ranks and applied LoRA targets in Table~\ref{tab:ablation_lora}.
We find that trainable parameters is essential for domain adaptation.
By increasing the LoRA rank to 256 with around 610M trainable parameters, the performance of parameter efficient fine-tuning can match full fine-tuning on domain adaption.

\section{Conclusion}
In this paper, we focus on improving the efficiency of domain adaptation through extended-text reading comprehension based on LLM and Clustering.
To achieve it, we propose a domain adaptation method that incorporates LLM-based data preprocessing, length-based clustering, and parameter-efficient domain adaptation.
For LLM-based data preprocessing, we improve regex-based patterns in AdaptLLM by exploiting the ability of LLMs to generate question-answer pairs to help the model learn domain-specific knowledge.
For length-based clustering, we further extend the context of question answering by concatenating similar documents into the same input.
For parameter-efficient domain adaptation, we argue that LoRA can be more efficient than full fine-tuning for domain adaptation with proper settings.
Experiments show that our method are effective to leverage unsupervised domain-specific corpus to improve the performance of domain-specific tasks.


\bibliography{custom}
\bibliographystyle{acl_natbib}

\end{document}